\documentclass{article} % For LaTeX2e
\usepackage{iclr2020_conference,times}

\usepackage{amsmath,amsfonts,bm}

\def\eqref#1{equation~\ref{#1}}
\def\1{\bm{1}}

\DeclareMathAlphabet{\mathsfit}{\encodingdefault}{\sfdefault}{m}{sl}
\SetMathAlphabet{\mathsfit}{bold}{\encodingdefault}{\sfdefault}{bx}{n}

\DeclareMathOperator*{\argmax}{arg\,max}
\DeclareMathOperator*{\argmin}{arg\,min}

\usepackage{wrapfig,lipsum,booktabs}
\usepackage{multirow}
\usepackage[colorlinks = true,
            linkcolor = blue,
            urlcolor  = blue,
            citecolor = black,
            anchorcolor = blue]{hyperref}
\usepackage{url}
\usepackage{wrapfig}
\usepackage{graphicx}
\usepackage{enumitem}
\usepackage{subfigure}

\title{Triple Wins: Boosting Accuracy, Robustness and Efficiency Together by Enabling Input-Adaptive Inference}

\author{Ting-Kuei Hu\thanks{Equal contribution}, \, Tianlong Chen\footnotemark[1], \, Haotao Wang \, Zhangyang Wang  \\
Department of Computer Science and Engineering\\
Texas A\&M University, USA\\
\texttt{\{tkhu,wiwjp619,htwang,atlaswang\}@tamu.edu} \\}

\newcommand{\bbR}{{\mathbb{R}}}
\newcommand{\bbD}{{\mathbb{D}}}

\newcommand{\eg}{e.g.,\ }
\newcommand{\ie}{i.e.,\ }

\newcommand{\changeurlcolor}[1]{\hypersetup{urlcolor=#1}}       
\iclrfinalcopy % Uncomment for camera-ready version, but NOT for submission.
\begin{document}

\maketitle

\begin{abstract}
Deep networks were recently suggested to face the odds between accuracy (on clean natural images) and robustness (on adversarially perturbed images) \citep{tsipras2019robustness}. Such a dilemma is shown to be rooted in the inherently higher sample complexity \citep{schmidt2018adversarially} and/or model capacity \citep{nakkiran2019adversarial}, for learning a high-accuracy and robust classifier. In view of that, give a classification task, growing the model capacity appears to help draw a \textit{win-win} between accuracy and robustness, yet at the expense of model size and latency, therefore posing challenges for resource-constrained applications. Is it possible to \textit{co-design} model accuracy, robustness and efficiency to achieve their \textit{triple wins}?

This paper studies multi-exit networks associated with input-adaptive efficient inference, showing their strong promise in achieving a ``sweet point'' in co-optimizing model accuracy, robustness and efficiency. Our proposed solution, dubbed \textit{Robust Dynamic Inference Networks} (RDI-Nets), allows for each input (either clean or adversarial) to adaptively choose one of the multiple output layers (early branches or the final one) to output its prediction. That multi-loss adaptivity adds new variations and flexibility to adversarial attacks and defenses, on which we present a systematical investigation. We show experimentally that by equipping existing backbones with such robust adaptive inference, the resulting RDI-Nets can achieve better accuracy and robustness, yet with over 30\%  computational savings, compared to the defended original models.

\end{abstract}

\vspace{-1.5em}
\section{Introduction}
\vspace{-1em}
Deep networks, despite their high predictive accuracy, are notoriously vulnerable to adversarial attacks \citep{goodfellow2014explaining,biggio2013evasion,szegedy2013intriguing,papernot2016limitations}. While many defense methods have been proposed to increase a model's \textit{robustness} to adversarial examples, they were typically observed to hamper its \textit{accuracy} on original clean images. \cite{tsipras2019robustness} first pointed out the inherent tension between the goals of adversarial robustness and standard accuracy in deep networks, whose provable existence was shown in a simplified setting. \cite{zhang2019theoretically} theoretically quantified the accuracy-robustness trade-off, in terms of the gap between the risk for adversarial examples versus the risk for non-adversarial examples. 

It is intriguing to consider whether and why the model accuracy and robustness have to be at odds. \cite{schmidt2018adversarially} demonstrated that the
number of samples needed to achieve adversarially robust generalization is polynomially larger than that needed for standard generalization, under the adversarial training setting. A similar conclusion was concurred by \cite{sun2019towards} in the standard training setting. \cite{tsipras2019robustness} considered the accuracy-robustness trade-off as an inherent trait of the data distribution itself, indicating that this phenomenon persists even in the limit of
infinite data. \cite{nakkiran2019adversarial} argued from a different perspective, that the complexity (e.g. capacity) of a robust classifier must be higher than that of a standard classifier. Therefore, replacing a larger-capacity classifier might effectively alleviate the trade-off. Overall, those existing works appear to suggest that, while accuracy and robustness are likely to trade off for a fixed classification model and on a given dataset, such trade-off might be effectively alleviated (``win-win''), if supplying more training data and/or replacing a larger-capacity classifier.

On a separate note, deep networks also face the pressing challenge to be deployed on resource-constrained platforms due to the prosperity of smart Internet-of-Things (IoT) devices. Many IoT applications naturally demand security and trustworthiness, \eg, biometrics and identity verification, but can only afford limited latency, memory and energy budget. Hereby we extend the question: \textit{can we achieve a triple-win, \ie, an accurate and robust classfier while keeping it efficient?} 

%Well-crafted perturbations are added to the original images to cause the model to make a confident but erroneous prediction. The so-called adversarial examples have gathered significant attention to the research community. Recent research findings demonstrate that adversarial robustness comes at odds with standard accuracy. The upper bound can be escalated by increasing model complexity and sample complexity \citep{tsipras2019robustness,Alhussein2018robustness}. 

%Several defense approaches, such as ensemble adversarial training, achieve promising adversarial robustness at the expense of larger network capacity. However, demands on short latency and high robustness are increasing as the emergence of Internet-of-Things (IoT) devices. In such systems, each edge device grabs ambient data via chip hardware sensors and analyze them by low-cost processors. Such systems usually require security promises to resist outer adversaries but can only function under limited resource budget. The large complexity of robust CNN models becomes a critical obstacle for the deployment of such systems. Although a few research efforts have been made in exploring the trading relationship between adversarial robustness and model compression \citep{lin2019defensive,gui2019adversarially,ye2019adversarial}, they keep a bare and brittle balance since over-sparsification is more likely to introduce more fragility.

This paper makes an attempt in providing a positive answer to the above question. Rather than proposing a specific design of robust light-weight models, we reduce the average computation loads by input-adaptive routing to achieve triple-win. To this end, we introduce the input-adaptive dynamic inference  \citep{teerapittayanon2016branchynet,wang2017skipnet}, an emerging efficient inference scheme in contrast to the (non-adaptive) model compression, to the adversarial defense field for the first time. Given any deep network backbone (\eg, ResNet, MobileNet), we first follow \citep{teerapittayanon2016branchynet} to augment it with multiple early-branch output layers in addition to the original final output. Each input, regardless of clean or adversarial samples, adaptively chooses which output layer to take for its own prediction. Therefore, a large portion of input inferences can be terminated early when the samples can already be inferred with high
confidence.

\begin{figure}[ht]
    \centering
    \includegraphics[width=0.8\linewidth]{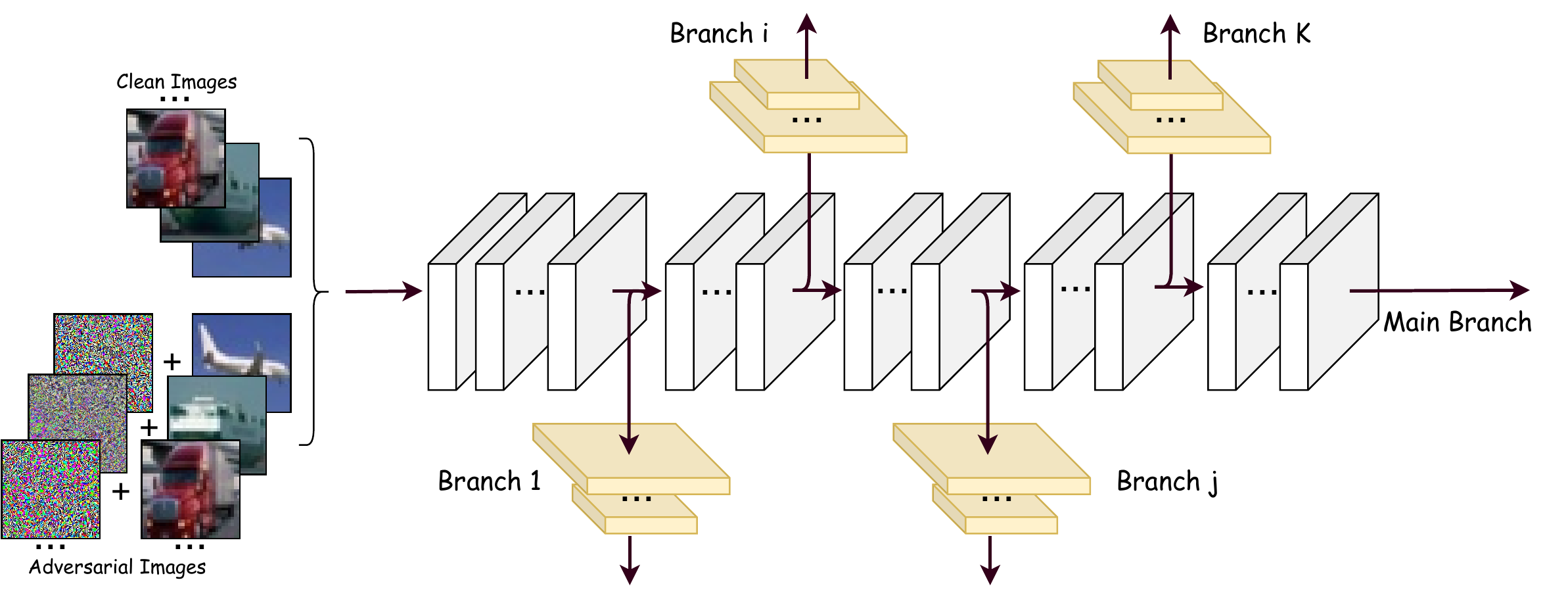}
    \vspace{-1em}
    \caption{Our proposed RDI-Net framework, a \textbf{defended} multi-output network enabling dynamic inference. Each image, being it clean or adversarially perturbed, adaptively picks one branch to exit.}
    \label{fig:branchy}
    \vspace{-1em}
\end{figure}

Up to our best knowledge, no existing work studied adversarial attacks and defenses for an adaptive multi-output model, as the multiple sources of losses provide much larger flexibility to compose attacks (and therefore defenses), compared to the typical single-loss backbone. We present a systematical exploration on how to (white-box) attack and defense our proposed multi-output network with adaptive inference, demonstrating that the composition of multiple-loss information is critical in making the attack/defense strong. Fig. \ref{fig:branchy} illustrates our proposed \textit{Robust Dynamic Inference Networks} (RDI-Nets). We show experimentally that the input-adaptive inference and multi-loss flexibility can be our friend in achieving the desired ``triple wins''. With our best defended RDI-Nets, we achieve better accuracy and robustness, yet with over 30\% inference computational savings, compared to the defended original models as well as existing solutions co-designing robustness and efficiency \citep{gui2019model,guo2018sparse}. The codes can be referenced from \changeurlcolor{blue}\href{https://github.com/TAMU-VITA/triple-wins}{\textit{https://github.com/TAMU-VITA/triple-wins.}}

%This work aim to bridge the gap between robustness and network efficiency by proposing a framework to construct robust and efficient dynamic inference networks (RDI-Nets). The resultant RDI-Nets can be interpreted as a special case of ensemble networks with weight-sharing structure and dynamic inference paths. RDI-Nets leverage the benefits of multi-output networks and dodging their deficiency on network efficiency by firstly augmenting appropriate adversarial images during adversarial defense followed by enabling dynamic inference mechanism during the testing stage. To be more specific, the former is achieved by exploiting multiple output sources to generate candidate adversarial samples and pick the most malicious one to defense, and the latter is carried out by inserting additional side branch classifiers throughout the network to make certain samples exit from the early branches.  As illustrated in Fig \ref{fig:branchy}, the resultant RDI-Nets contains multiple paths for the sample inference and impose difficulty on the attackers by exiting adversarial samples from an appropriate path.

\section{Related Work}
\vspace{-0.5em}
\subsection{Adversarial Defense}
\vspace{-0.5em}
A magnitude of defend approaches have been proposed \citep{kurakin2016adversarial,xu2017feature,song2017pixeldefend,liao2018defense}, although many were quickly evaded by new attacks \citep{carlini2017towards,baluja2017adversarial}. One
strong defense algorithm that has so far not been fully compromised is adversarial training \citep{madry2017towards}. It searches for adversarial images to augment the training procedure, although at the price of higher training costs (but not affecting inference efficiency). However, almost all existing attacks and defenses focus on a single-output classification (or other task) model. We are unaware of prior studies directly addressing attacks/defenses to more complicated networks with multiple possible outputs. 

One related row of works are to exploit model ensemble \citep{tramer2017ensemble,strauss2017ensemble} in adversarial training. The gains of the defended ensemble compared to a single model could be viewed as the benefits of either the benefits of diversity (generating stronger and more transferable perturbations), or the increasing model capacity (consider the ensembled multiple models as a compound one). Unfortunately, ensemble methods could amplify the inference complexity and be detrimental for efficiency. Besides, it is also known that injecting randomization at inference time helps mitigate adversarial effects \citep{xie2017mitigating,cohen2019certified}. Yet up to our best knowledge, no work has studied non-random, but rather input-dependent inference for defense. 

%Recent researches also indicate that adversarial robustness usually comes at odds with standard accuracy [\cite{tsipras2019robustness,Alhussein2018robustness}]. The upper bound of the trading relationship can be escalated by increasing the model complexity and sample complexity. Unfortunately, it also increases the difficulty of the deployment of complex deep convolutional neural networks on the resource-limited and latency-aware systems, such as IoT applications.
\vspace{-0.5em}
\subsection{Efficient Inference}
\vspace{-0.5em}

Research in improving deep network efficiency could be categorized into two streams: the \textit{static} way that designs compact models or compresses heavy models, while the compact/compressed models remain fixed for all inputs at inference; and the \textit{dynamic} way, that at inference the inputs can choose different computational paths adaptively, and the simpler inputs usually take less computation to make predictions. We briefly review the literature below.  

%Recent researches address the issue of network efficiency by exploring compact network components, seeking sparse networks and enabling input-adaptive inference. We provide a summarized review of the related literatures in the following.

\noindent \underline{\textit{Static: Compact Network Design and Model Compression.}} Many compact architectures have been specifically designed for resource-constrained applications, by adopting lightweight depthwise convolutions \citep{sandler2018mobilenetv2}, and group-wise convolutions with channel-shuffling \citep{ZhangZLS2018shufflenet}, to just name a few. For model compression, \cite{han2015learning} first proposed to sparsify deep models by removing non-significant synapses and then re-training to restore performance. Structured pruning was later on introduced for more hardware friendliness \citep{wen2016learning}. Layer factorization \citep{tai2015convolutional,yu2017compressing}, quantization \citep{wu2016quantized}, model distillation~\citep{wang2018adversarial} and weight sharing \citep{wu2018deep} have also been respectively found effective. 

%Research effors have been done on proposing lightweight depthwise convolutions [\cite{sandler2018mobilenetv2}], employing group-wise convolutions with channel-shuffling mechanism [\cite{ZhangZLS2018shufflenet}], learning resource-constrained structures by iteratively shrinking and expanding networks, to name a few.
%\vspace{-1em}
%\paragraph{Sparse Networks} 
%In [\cite{han2015learning}], the author sparsifies deep neural networks by thresholding non-significant synapses and re-training the network for performance recovery. Other recent works are proposed to jointly optimize network sparsity and network performance by penalizing the scale variables of batch normalization in the L$1$ norm sense [\cite{liu2017learning,huang2018data}].
%\vspace{-1em}
\noindent \underline{\textit{Dynamic: Input-Adaptive Inference.}} Higher inference efficiency could be also accomplished by enabling input-conditional execution. \cite{teerapittayanon2016branchynet,huang2018multiscale,kaya2019shallow} leveraged intermediate features to augment
multiple side branch classifiers to enable early predictions. Their methodology sets up the foundation for our work. Other efforts \citep{figurnov2017spatially,wang2017skipnet,wang2018energynet,wang2019dual} allow for an input to choose between passing through or skipping each layer. The approach could be integrated with RDI-Nets too, which we leave as future work.

%Other rearch efforts on conditional learning [\cite{panda2016conditional}], input-dependent skipping [\cite{wang2017skipnet}] also allow the samples to be conditionally processed based on their content.

\vspace{-0.5em}
\subsection{Bridging Robustness with Efficiency}
\vspace{-0.5em}
%Our work connects two fields: adversarial robustness and network efficiency. 

A few studies  recently try to link deep learning robustness and efficiency. \cite{guo2018sparse} observed that in a sparse deep network, appropriately sparsified weights improve robustness, whereas over-sparsification (e.g., less than 5\% nonzero weights) in turn makes the model more fragile. Two latest works \citep{ye2019adversarial,gui2019model} examined the robustness of compressed models, and concluded similar observations that the relationship between mode size and robustness depends on compression methods and are often non-monotonic. \cite{lin2019defensive} found that activation quantization may hurt robustness, but can be turned into effective defense if enforcing continuity constraints. 

%have conducted in this domain. \cite{ye2019second} exhibits there is an intrinsic relationship between adversarial robustness and network sparsity. Later on, \cite{ye2019adversarial,gui2019adversarially} demonstrate weight pruning sparsifying network while still preserving the adversarial robustness. Defensive Quantization from \cite{lin2019defensive} shows the quantization module correctly clarifies adversarial samples while maintaining the network efficiency. All of this works are seeking efficient and robust architectures to achieve high accuracy and robustness simultaneously.

Different from above methods that tackle robustness from \textit{static} compact/compressed models, the proposed RDI-Nets are the first to address robustness from the \textit{dynamic} input-adaptive inference. Our experiment results demonstrate the consistent superiority of RDI-Nets over those static methods (Section 4.3). Moreover, applying \textit{dynamic} inference top of those static methods may further boost the robustness and efficiency, which we leave as future work.

\vspace{-0.5em}
\section{Approach}
\vspace{-0.5em}
With the goal of achieving inference efficiency, we first look at the setting of multi-output networks and the specific design of RDI-Net in Section \ref{sec:3-1}. Then we define three forms of adversarial attacks for multi-output networks in Section \ref{sec:3-2} and their corresponding defense methods in Section \ref{sec:3-3}.

Note that RDI-Nets achieve ``triple wins''via reducing the average computation loads through input-adaptive routing. It is not to be confused with any specifically-designed robust light-weight model. 

\vspace{-0.5em}
\subsection{Designing RDI-Nets for Higher Inference Efficiency}\label{sec:3-1}
\vspace{-0.5em}
Given an input image $x$, an $N$-output network can produce a set of predictions $[\hat{y_{1}}, ..., \hat{y_{N}}]$ by a set of transformations $[f_{\theta_{1}}(\cdot), ..., f_{\theta_{N}}(\cdot)]$. $\theta_{i}$ denote the model parameter of $f_{\theta_{i}}$, $i = 1, ..., N$, and $f_{\theta_{i}}$s will typically share some weights. With an input $x$, one can express $\hat{y_{i}}=f_{\theta_{i}}(x)$. We assume that the final prediction will be \underline{one} \textbf{chosen} (NOT fused) from $[\hat{y_{1}}, ..., \hat{y_{N}}]$ via some \underline{deterministic} strategy.

We now look at RDI-Nets as a specific instance of multi-output networks, specifically designed for the goal of more efficient, input-adaptive inference. As shown in Fig. \ref{fig:branchy}, \textbf{for any deep network} (\eg, ResNet, MobileNet), we could append $K$ side branches (with negligible overhead) to allow for early-exit predictions. In other words, it becomes a $(K+1)$-output network, and the \textit{subneworks} with the $K+1$ exits, from the lowest to the highest (the original final output), correspond to $[f_{\theta_{1}}(\cdot), ..., f_{\theta_{K+1}}(\cdot)]$. They share their weights in a \textit{nested} fashion: $\theta_{1} \subseteq \theta_{2} ... \subseteq \theta_{K+1}$, with $\theta_{K+1}$ including the entire network's parameters.

% We now look at RDI-Nets as a specific instance of general multi-output networks, specifically designed for the goal of more efficient, input-adaptive inference. As shown in Fig. \ref{fig:branchy}, \textbf{for any deep network} (\eg, ResNet, MobileNet), we could append $K$ side branches (with negligible overhead) to allow for early-exit predictions. In other words, it becomes a $(K+1)$-output network, and the \textit{subneworks} with the $K+1$ exits, from the lowest to the highest (the original final output), correspond to $[f_{1}(\theta_{1}|\cdot), ..., f_{K+1}(\theta_{K+1}|\cdot)]$. They share their weights in a \textit{nested} fashion: $\theta_{1} \subseteq \theta_{2} ... \subseteq \theta_{K+1}$, with $\theta_{K+1}$ including the entire network's parameters. 

%$N$ losses (including the original final one), augment any deep network backbone

%Note that the $N$ functions will be nested in RDI-Nets

Our deterministic strategy in selecting one final output follows \citep{teerapittayanon2016branchynet}. We set a confidence threshold $t_k$ for each $k$-th exit, $k = 1, ..., K+1$, and each input $x$ will terminate inference and output its prediction in the earliest exit (smallest $k$), whose softmax entropy (as a confidence measure) falls below $t_k$. All computations after the $k$-th exit will not be activated for this $x$. Such a progressive and early-halting mechanism effectively saves unnecessary computation for most easier-to-classify samples, and applies in both training and inference. Note that, if efficiency is not the concern, instead of choosing (the earliest one), we could have designed an adaptive or randomized fusion of all $f_{\theta_{i}}$ predictions: but that falls beyond the goal of this work. 

The training objective for RDI-Nets could be written as
\begin{equation}
L_{RDI} = \sum_{i=1}^{K+1} w_{i}[(\phi(f_{i}(\theta_{i}|x), y) + \phi(f_{i}(\theta_{i}|x^{adv}), y)], \label{eqn:weighted-loss}
\end{equation}
For each exit loss, we minimize a hybrid loss of accuracy (on clean $x$) and robustness (on $x^{adv}$). The $K+1$ exits are balanced with a group of weights $\{w_{i}\}_{i=1}^{K+1}$. More details about RDI-Net structures, hyperparameters, and inference branch selection can be founded in Appendix \ref{sec:6-1}, \ref{sec:6-2}, and \ref{sec:6-3}.

In what follows, we discuss three ways to generate $x^{adv}$ in RDI-Nets, and then their defenses.

\vspace{-0.5em}
\subsection{Three Attack Forms on Multi-Output Networks}\label{sec:3-2} 
\vspace{-0.5em}
We consider white box attacks in this paper. Attackers have access to the model's parameters, and aim to generate an adversarial image $x^{adv}$ to fool the model by perturbing an input $x$ within a given magnitude bound. 

We next discuss three \textit{attack forms} for an $N$-output network. Note that they are independent of, and to be distinguished from \textit{attacker algorithms} (\eg, PGD, C\&W, FGSM): the former depicts the \textit{optimization formulation}, that can be solved any of the attacker algorithms. 

%For each $\hat{y_{i}}$, we could obtain it by the equation $\hat{y_{i}}=f_{i}(\theta_{i}|x)$, where $f_{i}(\theta_{i}|\cdot)$ denotes the $i$th transformation function, which maps $x$ to $\hat{y_{i}}$, and $\theta_{i}$ is the learning parameters with respect to $f_{i}(\theta_{i}|\cdot)$. 

%To study the adversarial robustness of a multi-output network, we consider common white-box attack setting. The white-box attack allows attackers to access to the model's parameters and aims to generate an adversarial images $x^{adv}$ to be misclassified by the model. To this end, the attacker perturbs the input image $x$ into $x^{adv}$ by adding a perturbation within a given bounded magnitude. Unlike previous well-studied attacks on single output network, three types of adversarial attack, \ie single attack, average attack, and max-average attack, arise due to the multi-output network setting.
\vspace{-0.5em}
\paragraph{Single Attack}
Naively extending from attacking single-output networks, a single attack is defined to maximally fool one $f_{\theta_{i}}(\cdot)$ only, expressed as: 
%the performance of $f_{i}(\theta_{i}|\cdot)$. Provided an input image $x$, an adversarial image $x^{adv}$ could be derived by the following equation,
\begin{equation}
\hspace{-0.01in} 
%x^{adv} \leftarrow 
x_{i}^{adv} = \argmax_{x'\in |x'-x|_{\infty}\leq\epsilon}|\phi(f_{\theta_{i}}(x'), y)|,\label{eqn:single-attack}
\end{equation}
where $y$ is the ground truth label, and $\phi$ is the loss for $f_{\theta_{i}}$ (we assume softmax for all). $\epsilon$ is the perturbation radius and we adopt $\ell_\infty$ ball for an empirically strong attacker. Naturally, an $N$-output network can have $N$ different single attacks. However, each single attack is derived without being aware of other parallel outputs. The found $x_{i}^{adv}$ is not necessarily transferable to other $f_{\theta_{j}}$s ($j \neq i$), and therefore can be easily bypassed if $x$ is re-routed through other outputs to make its prediction.

\vspace{-0.5em}
\paragraph{Average Attack}
Our second attack maximizes the average of all $f_{\theta_{i}}$ losses, so that the found $x^{adv}$ remains in effect no matter which one $f_{\theta_{i}}$ is chosen to output the prediction for $x$:
%can be effective to every transformation function. The desired $x^{adv}$ can be derived via the following equation,
\begin{equation}
\hspace{-0.01in} x_{avg}^{adv} = \argmax_{x'\in |x'-x|_{\infty}\leq\epsilon}|\frac{1}{N}\sum_{j=1}^{N}\phi(f_{\theta_{j}}(x'), y)|,\label{eqn:avg-attack}
\end{equation}
%\textbf{Eq.\ref{eqn:avg-attack}} depicts the optimization procedure that aims to search $x^{adv}$ by maximizing the average of the losses of every transformation function $f_{j}(\theta_{j}|\cdot)|_{j=1}^{N}$. 
The average attack addresses takes into account the attack transferablity and involves all $\theta_{j}$s into optimization. However, while only one output will be selected for each sample at inference, the average strategy might weaken the individual defense strength of each $f_{\theta_{i}}$.

%the issue of the transferability of model vulnerability by involving all the learnable parameters $\theta_{j}|_{j=1}^{N}$ into the maximization procedure, leading any of the transformation function that might be possibly selected during the inference to be vulnerable to $x^{adv}$ to some degrees. Nevertheless, the optimal performance degradation might not be able to be expected due to the fact that \textbf{Eq.\ref{eqn:avg-attack}} targets at maximizing average loss of every transformation function rather than maximizing the loss of any transformation function that could be possibly selected during the inference stage. Low success rate of attacks might occur due to this misalignment of loss maximization.

\vspace{-0.5em}
\paragraph{Max-Average Attack} Our third attack aims to emphasize individual output defense strength, more than simply maximizing an all-averaged loss. We first solve the $N$ single attacks $x^{adv}_{i}$ as described in Eqn. \ref{eqn:single-attack}, and denote their collection as $\Omega$. We then solve the max-average attack via the following:
\begin{equation}
\hspace{-0.01in} x_{max}^{adv} \leftarrow x_{i^*}^{adv} \mbox{, where } x^{adv}_{i^{*}}\in\Omega\mbox{ and } i^* = \arg\max_{i}|\frac{1}{N}\sum_{j=1}^{N}\phi(f_{\theta_{j}}(x_{i}^{adv}), y)|.\label{eqn:max-attack}
\end{equation}
%Rather than acquiring $x^{adv}$ by directly maximizing the average loss of transformation functions, we firstly nominates a set $\Omega$ containing candidate adversarial images. Then, the most malicious one in $\Omega$ is picked for the final inference. We exploit the property of multi-output networks to generate $\Omega$ by the following equation,

%\begin{equation}
%\hspace{-0.01in} x^{adv}_{i}\in\Omega \mbox{, where } x_{i}^{adv} = \argmax_{x'\in |x'-x|_{\infty}\leq\epsilon}|\phi(f_{i}(\theta_{i}|x'), y)|.\label{eqn:max-attack-2}
%\end{equation}
Note Eqn. \ref{eqn:max-attack} differs from Eqn. \ref{eqn:avg-attack} by adding an $\Omega$ constraint to balance between ``commodity" and ``specificity". The found $x_{max}^{adv}$ \textbf{both} strongly increases the averaged loss values from all $f_{i}$s (therefore possessing transferablity), \textbf{and} maximally fools one individual $f_{\theta_{i}}$s as it is selected from the collection $\Omega$ of single attacks. 

\vspace{-0.5em}
\subsection{Defence on Multi-Output Networks}\label{sec:3-3} 
\vspace{-0.5em}

For simplicity and fair comparison, we focus on adversarial training \citep{madry2017towards} as our defense framework, where the three above defined attack forms can be plugged-in to generate adversarial images to augment training, as follows ($\Theta$ is the union of learnable parameters):
%In order to reliably defend models that are robust to adversarial attacks, it is necessary to augment the adversarial images $x^{adv}$ during the training procedure. The first step towards adversarial robustness is to specify an attack model and one key question arises to the defense of multi-output networks is \textbf{"what adversarial images should be augmented during adversarial training?"} To well defend multi-output networks, we leverage the attack settings discussed in sec $\ref{sec:3-1}$ to define three types of defending methods, \ie single defense, average defense, and max-average defense. The corresponding produced $x^{adv}$ can be augmented during the training procedure, which can be described in the following equation,
\begin{equation}
\hspace{-0.01in} \theta_{i}\in\Theta \mbox{, where } \theta_{i} = \argmin_{\theta'}|\phi(f_{\theta_{i}}(x), y) + \phi(f_{\theta_{i}}(x^{adv}), y) |.\label{eqn:sur-2}, 
\end{equation}
where $x^{adv} \in \{x_{i}^{adv}, x_{avg}^{adv}, x_{max}^{adv}\}$. As $f_{i}$s partially share their weights $\theta_{i}$ in a multi-output network, the updates from different $f_{i}$s will be averaged on the shared parameters.

%When there is parameters Here, $\Theta$ is the union of learnable parameters of each transformation function, and the objective of adversarially training a multi-output network is to find the parameters $\theta_{i}$ that minimizing the empirical risk of each transformation function over $x$ and $x^{adv}$ under a suitable loss function $\phi$.
\vspace{-0.5em}

\vspace{-0.5em}
\section{Experimental Results}\label{sec:performance}
\vspace{-0.5em}
\subsection{Experimental Setup}
\vspace{-0.5em}
\paragraph{Evaluation Metrics}
We evaluate accuracy, robustness, and efficiency, using the metrics below: 
\vspace{-0.1in}
\begin{itemize}[leftmargin=15pt, itemsep=1pt]
\item Testing Accuracy (\textbf{TA}): the classification accuracy on the original clean test set.
\item Adversarial Testing Accuracy (\textbf{ATA}): Given an attacker, ATA stands for the classification accuracy on the attacked test set. It is the same as the ``robust accuracy" in \citep{zhang2019theoretically}. 
%The attack algorithm is firstly defined, and the adversarial sample for each sample in the original testing set is created according to the attack algorithm. ATA stands for the classification accuracy on the adversarial testing set. It can be interpreted as an indicator of the robustness of the model.
\item Mega Flops (\textbf{MFlops}): The number of million floating-point multiplication operations consumed on the inference, averaged over the entire testing set. 
%of the target testing set during the inference stage. For simplicity, we only consider the multiplication operations.
\end{itemize}
\vspace{-0.1in}

\vspace{-0.5em}

\paragraph{Datasets and Benchmark Models} We evaluate three representative CNN models on two popular datasets: SmallCNN on MNIST \citep{qi2018nipe}; ResNet-38 \citep{he2016resnet} and MobileNet-V2 \citep{sandler2018mobilenetv2} on CIFAR-10. The three networks span from simplest to more complicated, and covers a compact backbone. All three models are defended by adversarial training, constituting \textbf{strong baselines}. Table \ref{tab:benchmark} reports the models, datasets, the attacker algorithm used in attack \& defense, and thee TA/ATA/MFlops performance of three defended models.

%State-of-the-art CNN networks, \ie SmallCNN on MNIST [\cite{qi2018nipe}], and ResNet/MobileNet-V2 on CIFAR-10 [\cite{he2016resnet,mark2018mnv2}], are selected to serve as the backbone networks for the construction of dynamic inference networks. We benchmark the performance of the adversarial training of each CNN backbones and the details are summarized in table \ref{tab:benchmark}.

\vspace{-0.5em}
\paragraph{Attack and Defense on RDI-Nets} 
We build RDI-Nets by appending side branch outputs for each backbone. For SmallCNN, we add two side branches ($K = 2$). For ResNet-38 and MobileNet-V2, we have $K = 6$ and $K = 2$, respectively. The branches are designed to cause negligible overheads: more details of their structure and positions can be referenced in Appendix \ref{sec:6-2}. We call those result models RDI-SmallCNN, RDI-ResNet38 and RDI-MobileNetV2 hereinafter.

    We then generate attacks using our three defined forms. Each attack form could be solved with various attacker algorithms (\eg PGD, C\&W, FGSM), and by default we solve it with the same attacker used for each backbone in Table \ref{tab:benchmark}. If we fix one attacker algorithm (\eg PGD), then TA/ATA for a single-output network can be measured without ambiguity. Yet for ($K$+1)-output RDI-Nets, there could be at least $K$+3 different ATA numbers for one defended model, depending on what attack form in Section \ref{sec:3-1} to apply ($K$+1 single attacks, 1 average attack, and 1 max-average attack). For example, we denote by \textbf{ATA (Branch1)} the ATA number when applying the single attack generated from the first side output branch (\eg $x_{1}^{adv}$); similarly elsewhere.

We also defend RDI-Nets using adversarial training, using the forms of adversarial images to augment training. By default, we adopt three adversarial training defense schemes: \textbf{Main Branch} (single attack using $x_{K+1}^{adv}$)\footnote{We tried adversarial training using other $K$ earlier side branch single attacks, and found their TA/ATA to be much more deteriorated compared to the main branch one. We thus report this only for compactness.},\textbf{Average} (using $x_{avg}^{adv}$),and \textbf{Max-Average} (using $x_{avg}^{max}$), in addition to the undefended RDI-Nets (using standard training) denoted as \textbf{Standard}. 

We cross evaluate ATAs of different defenses and attacks, since an ideal defense shall protect against all possible attack forms. To faithfully indicate the actual robustness, we choose the \textit{lowest number} among all $K+3$ ATAs, denoted as \textbf{ATA (Worst-Case)}, as the robustness measure for an RDI-Net.

%and three types of adversarial samples are augmented during the defense training: ($1$) samples by maximizing main branch loss; ($2$) samples by maximizing average loss; and ($3$) samples by maximizing max-average loss. For the attack setting, we assume that the attacker can fully utilize every outputs of RDI-Nets. Thus, we enumerate all possible attack settings that attackers can adopt during the inference stage. We report the ATA of each attack setting, where we denote "\textbf{ATA} (Branch1)" as the setting that the attacker selects the branch$1$ loss as the source to generate adversarial samples, and so on.

\begin{table}[ht]
\centering
\vspace{-1.5em}
\caption{Benchmarking results of adverserial training of three networks. PGD-40 denotes running the projected gradient descent attacker \citep{madry2017towards} for 40 iterations. We set the perturbation size as $0.3$ for MNIST and $8/255$ for CIFAR-$10$ in $\ell_\infty$ norm (adopted by all following experiments).}
\vspace{0.1in}
\label{tab:benchmark}
\resizebox{0.7\textwidth}{!}{
\begin{tabular}{lllllll}
\hline
\textbf{Model} & \textbf{Dataset} & \textbf{Defend} & \textbf{Attack} & \textbf{TA} & \textbf{ATA} & \textbf{MFlops} \\ \hline
SmallCNN       & MNIST            & PGD-40          & PGD-40          & 99.49\%     & 96.31\%      & 9.25           \\
ResNet-38      & CIFAR-10         & PGD-10          & PGD-20          & 83.62\%     & 42.29\%      & 79.42          \\
MobileNetV2   & CIFAR-10         & PGD-10          & PGD-20          & 84.42\%     & 46.92\%      & 86.91    \\ \hline
\end{tabular}}
\vspace{-1em}
\end{table}

\vspace{-0.5em}
\subsection{Evaluation and Analysis}
\vspace{-0.5em}
\paragraph{MNIST Experiments}
%To construct a dynamic inference network, we firstly augment two branch classifiers on SmallCNN, and we denote this network as RDI-SmallCNN. The branch positions of this two extra classifiers are set to be equidistant through the network. 

The MNIST experimental results on RDI-SmallCNN are summarized in table \ref{tab:smallcnn-branch}, with several meaningful observations to be drawn. \underline{First}, the undefended models (Standard) are easily compromised by all attack forms. \underline{Second}, The single attack-defended model (Main Branch) achieves the best ATA against the same type of attack, \ie ATA (Main Branch), and also seems to boost the closest output branch's robustness, \ie ATA (Branch 2). However, its defense effect on the further-away Branch 1 is degraded, and also shows to be fragile under two stronger attacks (Average, and Max-Average). \underline{Third}, both Average and Max-Average defenses achieve good TAs, as well as ATAs against all attack forms (and therefore Worst-Case), with Max-Average slightly better at both (the margins are small due to the data/task simplicity; see next two). 

Moreover, compared to the strong baseline of SmallCNN defended by PGD (40 iterations)-based adversarial training, RDI-SmallCNN with Max-Average defense wins in terms of both TA and ATA. Impressively, that comes together with 34.30\% computational savings compared to the baseline. Here the different defense forms do not appear to alter the inference efficiency much: they all save around 34\% - 36\% MFlops compared to the backbone. 
%A ``tripe win'' is thus achieved on RDI-SmallCNN. 

\vspace{-0.15in}
\begin{table}[ht]
\centering
\caption{The performance of RDI-SmallCNN. The "\textbf{Average MFlops}" is calculated by averaging the total flop costs consumed over the inference of the entire set (different samples take different FLOPs due to input-adaptive inference). The perturbation size and step size are $0.3$ and $0.01$, respectively.} 
\vspace{0.1in}
\label{tab:smallcnn-branch}
\resizebox{0.8\textwidth}{!}{
\begin{tabular}{|l|l|l|l|l|}
\hline
Defense Method & Standard  & Main Branch & Average  & Max-Average  \\ \hline
\hline
\textbf{TA} & 99.48\% & 99.50\% & 99.51\% & \textbf{99.52\%} \\ \hline
\hline
ATA (Branch 1) & 6.60\% & 60.50\% & 98.69\% & 98.52\% \\ \hline
ATA (Branch 2) & 3.16\% & 98.14\% & 97.64\% & 97.62\% \\ \hline
ATA (Main Branch) & 1.32\% & 96.70\% & 96.30\% & 96.43\% \\ \hline
ATA (Average) & 2.61\% & 61.35\% & 97.37\% & 97.42\% \\ \hline
ATA (Max-Average) & 2.10\% & 61.83\% & 96.82\% & 96.89\% \\ \hline
\textbf{ATA (Worst-Case)} & 1.32\% & 60.50\% & 96.30\% & \textbf{96.43\%} \\ \hline
\hline
Average MFlops & 5.89 & 5.89 & 5.95 & 6.08 \\ \hline
\textbf{Computation Saving} & 36.40\% & 36.40\% & 35.70\% & 34.30\% \\ \hline
\end{tabular}
}
\vspace{-0.5em}
\end{table}

%We compare the experimental results of three different proposed defense strategies. We observe that a well-defended DI-SmallCNN, \ie the one defended by "Max-Average Loss", can achieve highest \textbf{ATA} on the worst-cast attack. Although the model defended by "Main Branch Loss" achieves highest \textbf{ATA} on main branch attack, it suffers from significant performance drop on the other types of attacks. The rationale behind is that the model defended by the "Main Branch Loss" cannot resist to the vulnerability transferred by other outputs, leaving the attacker chances to fail the model by leveraging the losses from other outputs. The other two types of defense strategies perform well on most of attack settings since both of them consider the transferability of vulnerability during the optimization of the training. It is worth noting that, compared to original SmallCNN, the attacker cannot achieve higher ATA on the model defended by the "Max-Average Loss" even though attacker have adopted the worst-cast attack on it. The results indicate that our well-defended DI-SmallCNNs enjoy less computational usage and achieve more accurate and robust performance. 

\vspace{-0.5em}
\paragraph{CIFAR-10 Experiments}\label{sec:4-2} 
The results on RDI-ResNet38 and RDI-MobileNetV2 are presented in Tables \ref{tab:bnet38-branch} and \ref{tab:mnv2-branch}, respectively. Most findings seem to concur with MNIST experiments. \underline{Specifically}, on the more complicated CIFAR-10 classification task, Max-Average defense achieves much more obvious margins over Average defense, in terms of ATA (Worst-Case): 2.79\% for RDI-ResNet38, and 1.06\% for RDI-MobileNetV2. Interestingly, the Average defense is not even the strongest in defending average attacks, as Max-Average defense can achieve higher ATA (Average) in both cases. We conjecture that averaging all branch losses might ``over-smooth'' and diminish useful gradients. 

Compared to the defended ResNet-38 and MobileNet-V2 backbones, RDI-Nets with Max-Average defense achieve higher TAs and ATAs for both. Especially, the ATA (Worst-Case) of RDI-ResNet-38 surpasses the ATA of ResNet-38 defended by PGD-adversarial training by \textbf{1.03\%}, while saving around \textbf{30\%} inference budget. We find that different defenses on CIFAR-10 have more notable impacts on computational saving. Seemingly, a stronger defense (Max-Average) requires inputs to go through the scrutiny of more layers on average, before outputting  confident enough predictions: a sensible observation as we expect.

\begin{figure}[ht]
\begin{tabular}{ccc}
\subfigure[]{
\centering
   \includegraphics[height=0.14\textheight]{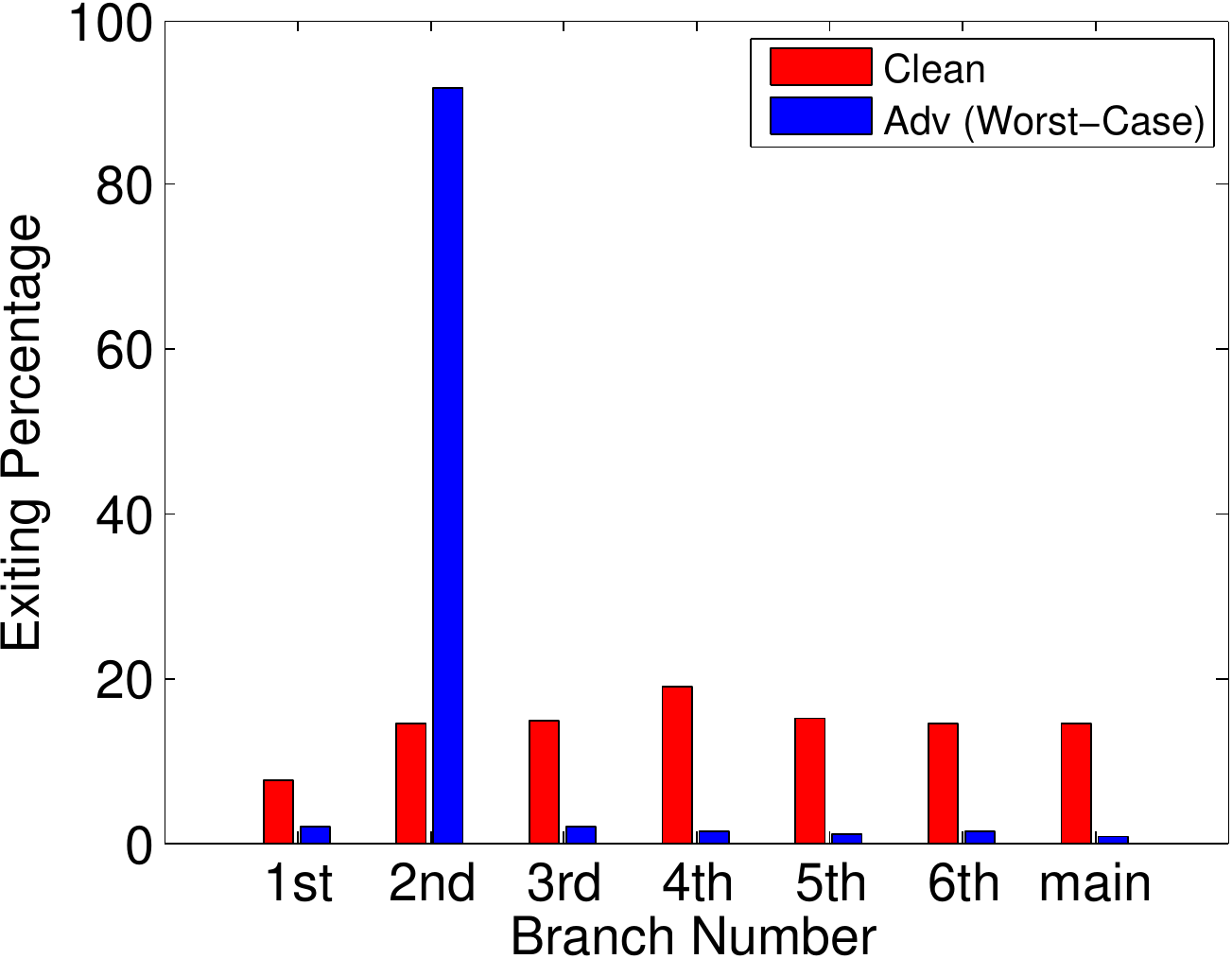}
   \label{fig:exit_main}
}
\quad
\centering
\subfigure[]{
   \includegraphics[height=0.14\textheight]{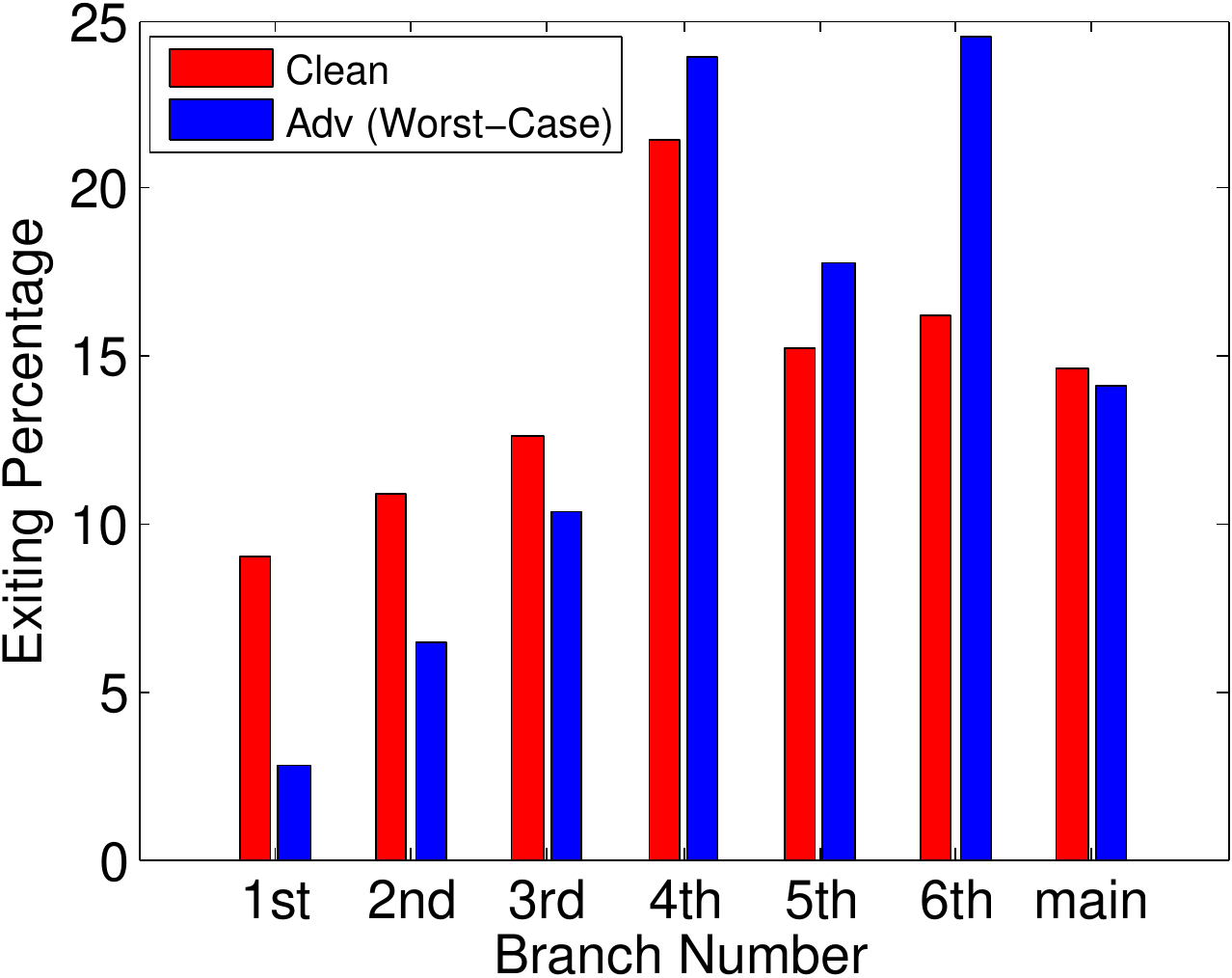}
   \label{fig:exit_avg}
}
\quad
\subfigure[]{
\centering
   \includegraphics[height=0.14\textheight]{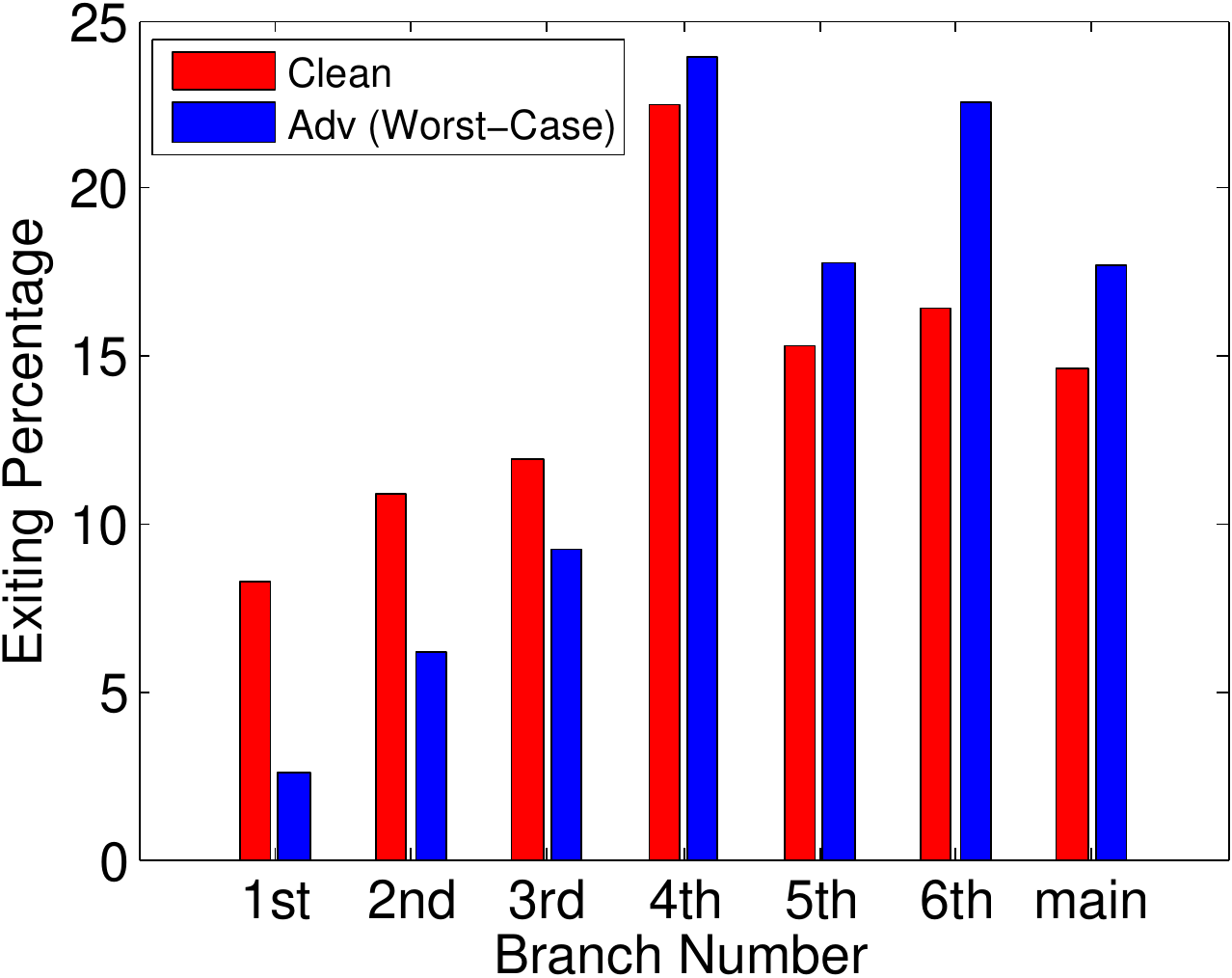}
   \label{fig:exit_max_avg}
}
\end{tabular}
\label{fig:exit_behavior}
\vspace{-0.2in}
\caption{ The exiting behaviours of RDI-ResNet38 defended by (a) Single attack defense (Main Branch); (b) Average defense; and (c) Max-Average defense.}
\vspace{-1em}
\end{figure}

\vspace{-0.5em}
\paragraph{Visualization of Adaptive Inference Behaviors} We visualize the exiting behaviors of RDI-ResNet38 in Fig \ref{fig:exit_behavior}. We plot each branch exiting percentage on clean set and adversarial sets (worst-case) of examples. A few interesting observations can be found. \underline{First}, we observe that the single-attack defended model can be easily fooled as adversarial examples can be routed through other less-defended outputs (due to the limited transferability of attacks between different outputs). \underline{Second}, the two stronger defenses (Average and Max-Average) show much more uniform usage of multiple outputs. Their routing behaviors for clean examples are almost identical. For adversarial examples, Max-Average tends to call upon the full inference more often (i.e., more ``conservative''). 

%by the vulnerability transferred from other outputs. Attacker can target at a non-well defended branch ( Branch 1 and Branch 2 in the Main-Branch defended model) to produce highly confident but erroneous predictions, resulting in disastrous performance drop. \underline{Second}, both Average and Max-Average defended models behave in a similar way, where the stronger defended model demands adversarial images to go through more scrutiny of layers than clean images on average.

%Two backbone architectures, \ie ResNet-38 and MobilenetV2, are considered in the experiments. The former is known by its excellent expressive power, and the latter is featured by its compact and efficient network components. To build dynamic inference networks on both backbone networks, we augment $6$ additional branch classifiers on ResNet-38 and $2$ extra branch classifiers on MobileNetV2. We denote RDI-ResNet38 and RDI-MobileNetV2 as the resultant networks. During the inference stage, we carefully adjust the uncertainty thresholds to make the contribution of those branches in the middle slightly larger than the branches in other positions. The experimental results are provided in the table \ref{tab:bnet38-branch} and the table \ref{tab:mnv2-branch}.
\begin{table}[ht]
\centering
\vspace{-1em}
\caption{The performance evaluation on RDI-ResNet38. The perturbation size and step size are $8/255$ and $2/255$, respectively.} 
\vspace{0.1in}
\label{tab:bnet38-branch}
\resizebox{0.8\textwidth}{!}{
\begin{tabular}{|l|l|l|l|l|}
\hline
Defence Method & Standard & Main Branch & Average & Max-Average \\ \hline
\hline
\textbf{TA} & 92.43\% & 83.74\% & 82.42\% & \textbf{83.79\%} \\ \hline
\hline
ATA (Branch1) & 0.12\% & 12.02\% & 71.56\% & 69.71\% \\ \hline
ATA (Branch2) & 0.01\% & 5.58\% & 66.67\% & 63.11\% \\ \hline
ATA (Branch3) & 0.04\% & 42.73\% & 60.65\% & 60.72\% \\ \hline
ATA (Branch4) & 0.06\% & 34.95\% & 50.17\% & 47.82\% \\ \hline
ATA (Branch5) & 0.06\% & 41.77\% & 44.83\% & 45.53\% \\ \hline
ATA (Branch6) & 0.11\% & 41.68\% & 45.83\% & 44.12\% \\ \hline
ATA (Main Branch) & 0.13\% & 42.74\% & 47.52\% & 49.82\% \\ \hline
ATA (Average) & 0.01\% & 9.14\% & 42.09\% & 43.32\% \\ \hline
ATA (Max-Average) & 0.01\% & 7.15\% & 40.53\% & 43.43\% \\ \hline
\textbf{ATA (Worst-Case)} & 0.01\% & 5.58\% & 40.53\% & \textbf{43.32\%} \\ \hline
\hline
Average MFlops & 29.41 & 48.27 & 56.90 & 57.81 \\ \hline
\textbf{Computation Saving} & 62.96\% & 39.20\% & 28.35\% & 27.20\% \\ \hline
\end{tabular}}
\vspace{-1em}
\end{table}

\begin{table}[ht]
\centering
\caption{The performance evaluation on RDI-MobilenetV2. The perturbation size and step size are $8/255$ and $2/255$, respectively.} 
\vspace{0.1in}
\label{tab:mnv2-branch}
\resizebox{0.8\textwidth}{!}{
\begin{tabular}{|l|l|l|l|l|}
\hline
Defence Method & Standard & Main Branch & Average & Max-Average \\ \hline
\hline
\textbf{TA} & 93.22\% & \textbf{85.28\%} & 82.14\% & 84.91\% \\ \hline
\hline
ATA (Branch1) & 0.35\% & 37.40\% & 67.65\% & 71.78\% \\ \hline
ATA (Branch2) & 0\% & 47.35\% & 50.38\% & 50.15\% \\ \hline
ATA (Main Branch) & 0\% & 46.69\% & 49.33\% & 46.99\% \\ \hline
ATA (Average) & 0\% & 35.20\% & 45.93\% & 47.00\% \\ \hline
ATA (Max-Average) & 0\% & 36.66\% & 49.33\% & 50.18\% \\ \hline
\textbf{ATA (Worst-Case)} & 0\% & 35.20\% & 45.93\% & \textbf{46.99\%} \\ \hline
\hline
Average MFlops & 49.78 & 52.81 & 58.23 & 60.84 \\ \hline
\textbf{Computation Saving} & 42.72\% & 39.23\% & 33.00\% & 29.99\% \\ \hline
\end{tabular}}
\vspace{-0.5em}
\end{table}

%Similar trends can be observed in both experiments of RDI-ResNet38 and RDI-MobileNetV2. Firstly, the models defended by "Main Branch Loss" achieves highest TA but are more likely to suffer from the vulnerability that is transferred from the other outputs. Secondly, the models defended by "Average Loss" and "Max-Average Loss" are more robust in the sense that the attacker can be hardly achieve higher success rate by maximizing an appropriate loss. Thirdly, the model defended by "Max-Average Loss" can achieve highest ATA on the worst-case attack with even improvement on the testing accuracy on RDI-MobileNetV2.

\vspace{-0.5em}
\subsection{Comparison with Defended Sparse Networks}
\vspace{-0.5em}

% A parallel strategy to achieve both robustness and efficiency, as suggested by \citep{guo2018sparse}, is to appropriately sparsify (prune) the model. 
% Inspired by \citep{ye2019second,gui2019adversarially}, we craft an alternative solution: first pruning ResNet-38 with as state-of-the-arts channel pruning technique \citep{huang2018spase}, and then defending the pruned network using the PGD-10 adversarial training. 

An alternative to achieve accuracy-robust-efficiency trade-off is by defending a sparse or compressed model.
Inspired by \citep{guo2018sparse,gui2019model}, we compare RDI-Net with Max-Average defense to the following baseline: first compressing the network with a state-of-the-art model compression method~\citep{huang2018spase}, and then defend the compressed network using the PGD-10 adversarial training. 
%In practice, we utilize the state-of-the-art channel pruning algorithm SSS~\citep{huang2018spase} as the model compression method in AP.
We sample different sparsity ratios in~\citep{huang2018spase} to obtain models of different complexities. 
%experiment with a series sparsity constraint trade-off parameter $\gamma$, which controls the sparsity level of the network, in order to obtain models with different complexity for AP. 
% We compare RDI-ResNet-38 with Max-Average defense with the AP. 
% We compare RDI-Net with Max-Average defense with the AP. 
% To be thorough, we vary the pruning ratio (\textcolor{red}{explain r and the numbers!!!!}) in \citep{huang2018spase}, as well as the confidence thresholds in RDI-ResNet-38, so that we could have multiple model results of different complexities for both methods.
Fig. \ref{fig:comparision} in Appendix visualizes the comparison on ResNet-38: for either method, we sample a few models of different MFLOPs. At similar inference costs (e.g., 49.38M for pruning + defense, and 48.35M for RDI-Nets), our proposed approach consistently achieves higher ATAs ($>$ 2\%) than the strong pruning + defense baseline, with higher TAs.

%where the size each marker (representing a resulting model) is proportional to the average MFLOPs at inference. One can observe that at similar inference costs (e.g., 49.38M for pruning + defense, and 48.35M for RDI-Nets), our proposed approach consistently achieves higher ATAs ($>$ 2\%) than pruning + defense, with higher TAs.}

\begin{wraptable}{r}{7cm}
\vspace{-2em}
\caption{Performance comparison between RDI-ResNet38 and ATMC.}\label{tab:atmc}
\vspace{-0.5em}
\begin{center}
\resizebox{0.5\textwidth}{!}{
\begin{tabular}{lccc}
\multicolumn{1}{c}{\bf Methods} & \multicolumn{1}{c}{\bf TA} & \multicolumn{1}{c}{\bf ATA} & \multicolumn{1}{c}{\bf MFlops} 
\\ \hline
ATMC (\cite{gui2019model}) & \textbf{83.81} & 43.02 & 56.82 \\
RDI-ResNet-38 (\textbf{Worst-Case}) & 83.79 & \textbf{43.32} & 57.81 \\
\end{tabular}}
\end{center}
\vspace{-1em}
\end{wraptable} 

We also compare with the latest ATMC algorithm \citep{gui2019model} that jointly optimizes robustness and efficiency, applied the same ResNet-38 backbone. As shown in Table~\ref{tab:atmc}, at comparable MFlops, RDI-ResNet-38 surpasses ATMC by 0.3\% in terms of ATA, with a similar TA.

\vspace{-0.5em}
\subsection{Generalized Robustness Against Other Attackers}
\vspace{-0.5em}
In the aforementioned experiments, we have only evaluated on RDI-Nets against ``deterministic" PGD-based adversarial images. We show that RDI-Nets also achieve better generalized robustness against other ``randomized" or unseen attackers. We create the new ``random attack": that attack will randomly combine the multi-exit losses, and summarize the results in Table \ref{tab:random}. We also follow the similar setting in \cite{gui2019model} and report the results against FGSM \citep{goodfellow2014explaining} and WRM \citep{simha2018wrm} attacker, in Tables \ref{tab:fgsm} and \ref{tab:wrm} respectively (more complete results can be found in Appendix \ref{sec:6-4}).

\begin{table}[ht]
\centering
\vspace{-0.5em}
\caption{Performance on RDI-ResNet38 against random attack. The perturbation size and step size are $8/255$ and $2/255$, respectively. More details of random attack can be referenced in Appendix \ref{sec:6-4}.}
\vspace{0.1in}
\label{tab:random}
\resizebox{0.8\textwidth}{!}{
\begin{tabular}{|l|l|l|l|l|}
\hline
Defence Method & Standard & Main Branch & Average & Max-Average \\ \hline
\hline
\textbf{TA} & 92.43\% & 83.74\% & 82.42\% & \textbf{83.79\%} \\ \hline
\hline
ATA (Random) & 0.01\% & 10.33\% & 43.11\% & \textbf{44.86}\% \\ \hline 
\hline
Average MFlops & 27.33 & 52.36 & 55.21 & 56.54 \\ \hline
\textbf{Computation Saving} & 65.58\% & 34.07\% & 30.48\% & 28.80\% \\ \hline
\end{tabular}}
\vspace{-0.8em}
\end{table}

\begin{table}[ht]
\centering
\vspace{-0.5em}
\caption{Performance on RDI-ResNet38 (defended with PGD) against FGSM attack (perturbation size is $8/255$). The original defended ResNet38 by PGD under the same attack has ATA $51.11\%$.}
\vspace{0.1in}
\label{tab:fgsm}
\resizebox{0.8\textwidth}{!}{
\begin{tabular}{|l|l|l|l|l|}
\hline
Defence Method & Standard & Main Branch & Average & Max-Average \\ \hline
\hline
\textbf{TA} & 92.43\% & 83.74\% & 82.42\% & \textbf{83.79\%} \\ \hline
\hline
ATA (Main Branch) & 11.51\% & 51.45\% & 53.64\% & 54.72\% \\ \hline
ATA (Average) & 11.41\% & 50.21\% & 51.81\% & 53.20\% \\ \hline
ATA (Max-Average) & 2.09\% & 47.53\% & 50.63\% & 52.40\% \\ \hline
\textbf{ATA (Worst-Case)} & 2.09\% & 47.53\% & 50.63\% & \textbf{51.05\%} \\ \hline
\hline
Average MFlops & 65.74 & 55.27 & 58.27 & 59.67 \\ \hline
\textbf{Computation Saving} & 17.21\% & 30.40\% & 26.40\% & 24.86\% \\ \hline
\end{tabular}}
\vspace{-0.8em}
\end{table}

\begin{table}[!ht]
\centering
\vspace{-0.5em}
\caption{Performance on RDI-ResNet38 (defended with PGD) against WRM attack (perturbation size is $0.3$). The original defended ResNet38 by PGD under the same attack has ATA $83.35\%$.}
\vspace{0.1in}
\label{tab:wrm}
\resizebox{0.8\textwidth}{!}{
\begin{tabular}{|l|l|l|l|l|}
\hline
Defence Method & Standard & Main Branch & Average & Max-Average \\ \hline
\hline
\textbf{TA} & 92.43\% & 83.74\% & 82.42\% & \textbf{83.79\%} \\ \hline
\hline
ATA (Main Branch) & 34.42\% & 83.74\% & 82.42\% & 83.78\% \\ \hline
ATA (Average) & 26.48\% & 83.69\% & 82.36\% & 83.77\% \\ \hline
ATA (Max-Average) & 23.51\% & 83.73\% & 82.40\% & 83.78\% \\ \hline
\textbf{ATA (Worst-Case)} & 23.51\% & 83.69\% & 82.36\% & \textbf{83.77\%} \\ \hline
\hline
Average MFlops & 50.05 & 50.46 & 52.89 & 52.38 \\ \hline
\textbf{Computation Saving} & 36.98\% & 36.46\% & 33.40\% & 34.04\% \\ \hline
\end{tabular}}
\vspace{-0.8em}
\end{table}

\vspace{-1em}
\section{Discussion and Analysis}
\vspace{-0.5em}
\paragraph{Intuition: Multi-Output Networks as Special Ensembles} Our intuition on defending multi-output networks arises from the success of ensemble defense in improving both accuracy and robustness \citep{tramer2017ensemble,strauss2017ensemble}, which also aligns with the model capacity hypothesis \citep{nakkiran2019adversarial}. A general multi-output network \citep{xu2019survey} could be decomposed by an ensemble of single-output models, with weight re-using enforced among them. It is thus more compact than an ensemble of independent models, and the extent of sharing weight calibrates ensemble diversity versus efficiency. Therefore, we expect a defended multi-output network to (mostly) inherit the strong accuracy/robustness of ensemble defense, while keeping the inference cost lower. 
\vspace{-0.5em}
\paragraph{Do "Triple Wins" Go Against the Model Capacity Needs?} We point out that our seemingly ``free'' efficiency gains (\eg not sacrificing TA/ATA) do not go against the current belief that a more accurate and robust classifier relies on a larger model capacity \citep{nakkiran2019adversarial}. From the visualization, there remains to be a portion of clean/adversarial examples that have to utilize the full inference to predict well. In other words, the full model capacity is still necessary to achieve our current TAs/ATAs. Meanwhile, just like in standard classification \citep{wang2017skipnet}, not all adversarial examples are born equally. Many of them can be predicted using fewer inference costs (taking earlier exits). Therefore, RDI-Nets reduces the ``effective model capacity'' averaged on all testing samples for overall higher inference efficiency, while not altering the full model capacity.

\vspace{-1em}
\section{Conclusion}
\vspace{-0.5em}
This paper targets to simultaneously achieve high accuracy and robustness and meanwhile keeping inference costs lower. We introduce the multi-output network and input-adaptive dynamic inference, as a strong tool to the adversarial defense field for the first time. Our RDI-Nets achieve the ``triple wins'' of better accuracy, stronger robustness, and around 30\% inference computational savings. Our future work will extend RDI-Nets to more dynamic inference mechanisms.

%We propose an appropriate defense method for RDI-Networks by leveraging the property of multi-output networks. Our extensive experiments have justified the efficacy of the proposed RDI-Networks by observing the following: (1) the attacker cannot achieve higher success rate no matter what attack strategies are adopted; (2) the well defended RDI-Networks can benefit from the dynamic inference procedure and enjoy less computational usage; (3) the resultant networks can remain same adversarial robustness without losing the accuracy on the original testing set.}

\vspace{-1em}
\section{Acknowledgement}
\vspace{-0.5em}
We would like to thank Dr. Yang Yang from Walmart Technology for highly helpful discussions throughout this project.
\bibliography{references}
\bibliographystyle{iclr2020_conference}
\appendix
\section{Learning Details of RDI-Nets}\label{sec:6-1} 
\paragraph{MNIST}
We adopt the network architecture from \citep{qi2018nipe} with four convolutions and three full-connected layers. We train for $13100$ iterations with a batch size of $256$. The learning rate is initialized as $0.033$ and is lowered by $10$ at $12000$th and $12900$th iteration. For hybrid loss, the weights $\{w_{i}\}_{i=1}^{N+1}$ are set as $\{1, 1, 1\}$ for simplicity. For adversarial defense/attack, we perform 40-steps PGD for both defense and evaluation. The perturbation size and step size are set as $0.3$ and $0.01$.
\paragraph{CIFAR-10}
We take ResNet-38 and MobileNetV2 as the backbone architectures. For RDI-ResNet38, we initialize learning rate as $0.1$ and decay it by a factor of 10 at $32000$th and $48000$th iteration. The learning procedure stops at $55000$ iteration. For RDI-MobileNetV2, the learning rate is set to $0.05$ and is lowered by $10$ times at $62000$th and $70000$th iteration. We stop the learning procedure at $76000$ iteration. For hybrid loss, we follow the discussion in \citep{hu2019anytime} and set $\{w_{i}\}_{i=1}^{N+1}$ of RDI-ResNet38 and RDI-MobileNetV2 as $\{0.5, 0.5, 0.7, 0.7, 0.9, 0.9, 2\}$ and $\{0.5, 0.5, 1\}$, respectively. For adversarial defense/attack, the perturbation size and step size are set as $8/255$ and $2/255$. 10-steps PGD is performed for defense and 20-steps PGD is utilized for evaluation.

\section{Network Structure of RDI-Nets}\label{sec:6-2} 
\begin{figure}[ht]
    \centering
    \includegraphics[width=0.7\linewidth]{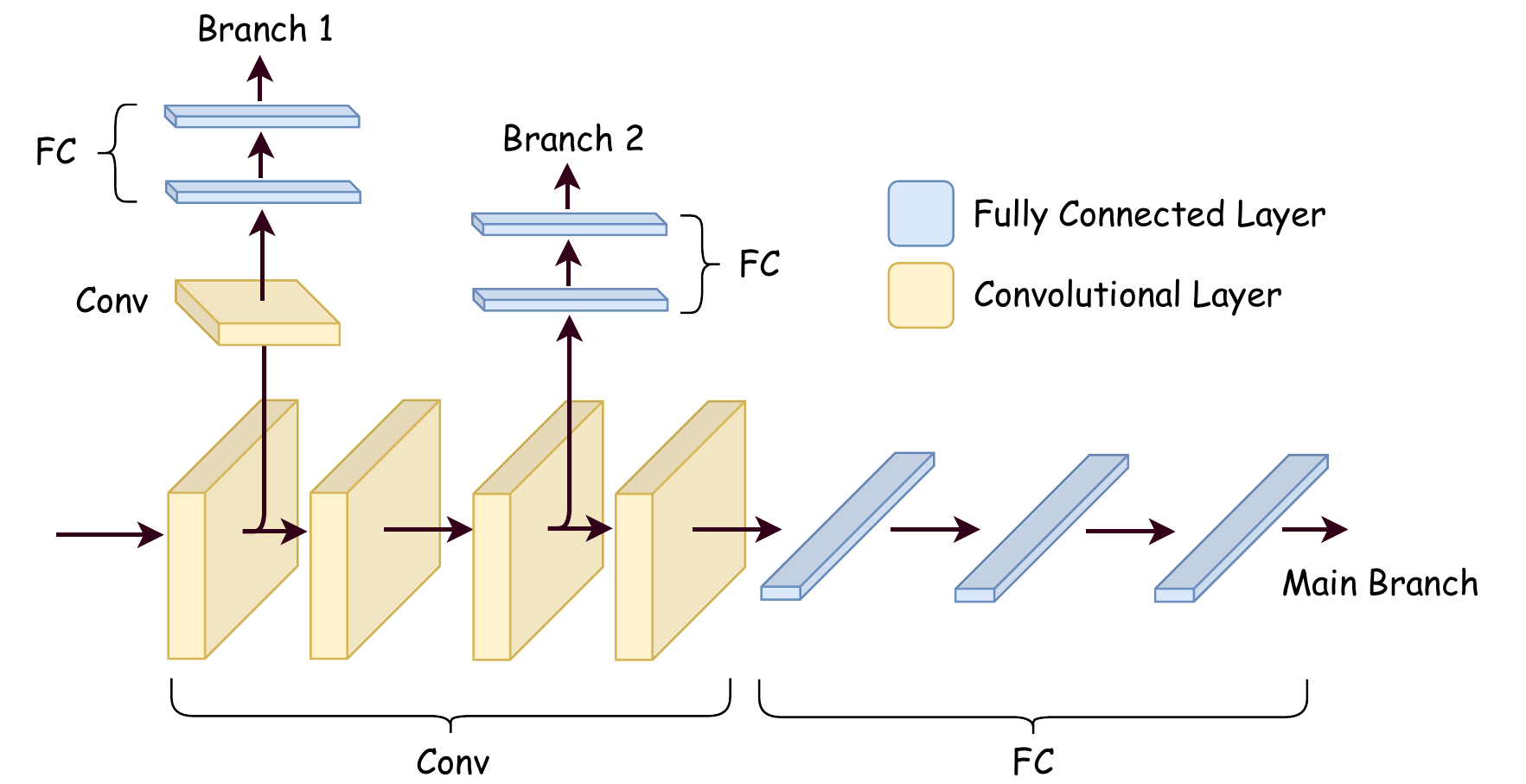}
    %\vspace{-1em}
    \caption{Network architecture of RDI-SmallCNN. Two branch classifiers are inserted after $1st$ convolutional layer and $3$rd convolutional layer in the original SmallCNN.}
    \label{fig:smallCNN}
    %\vspace{-1em}
\end{figure}

\begin{figure}[ht]
    \centering
    \includegraphics[width=1\linewidth]{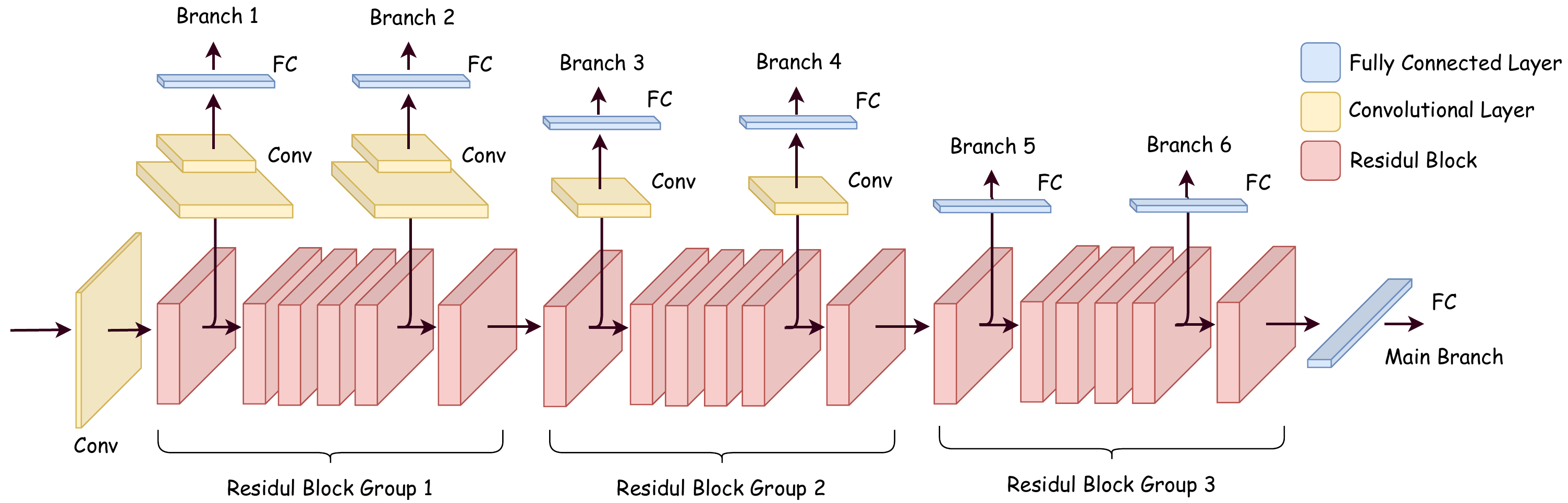}
    \vspace{-1em}
    \caption{Network architecture of RDI-ResNet38. In each residual block group, two branch classifiers are inserted after $1st$ residual block and $4$th residual block.}
    \label{fig:res38}
    \vspace{-1em}
\end{figure}

\begin{figure}[ht]
    \centering
    \includegraphics[width=1\linewidth]{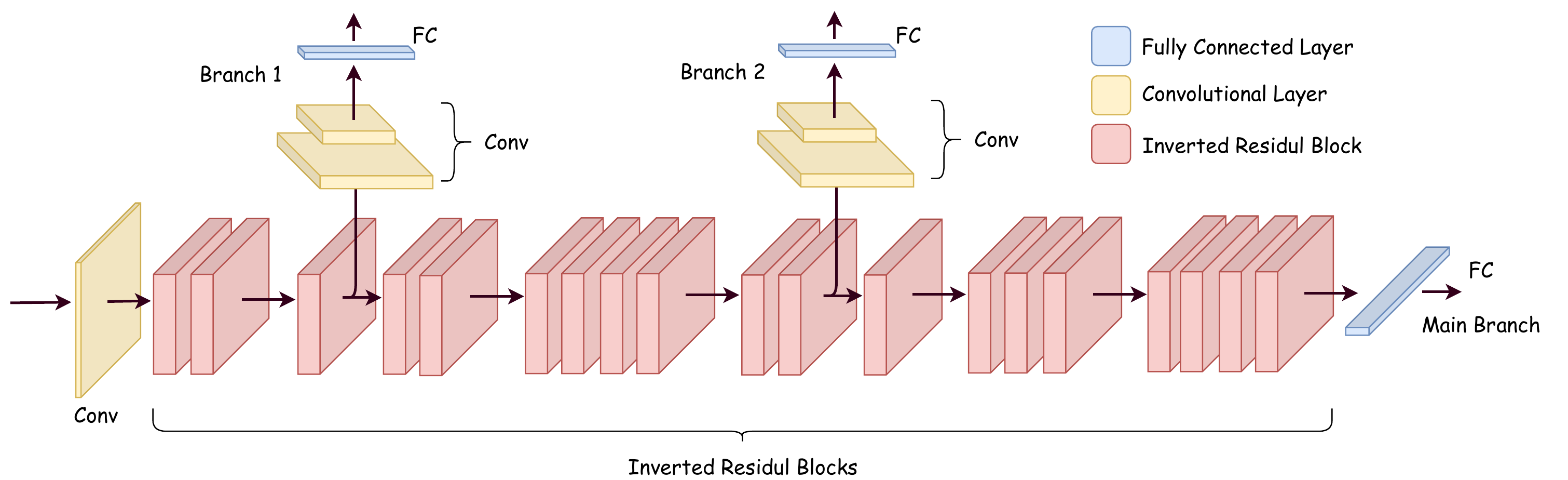}
    \vspace{-1em}
    \caption{Network architecture of RDI-MobilenetV2. Two branch classfiers are inserted after $3rd$ inverted residual block and $11$th inverted residual block in the orignal MobilenetV2.}
    \label{fig:mob}
    \vspace{-1em}
\end{figure}

To build RDI-Nets, we follow the similar setting in \cite{teerapittayanon2016branchynet} by appending additional branch classifiers at equidistant points throughout a given network, as illustrated in Fig \ref{fig:smallCNN}, Fig \ref{fig:res38} and Fig \ref{fig:mob}. A few pooling operations, light-weight convolutions and fully-connected layers are appended to each branch classifiers. Note that the extra flops introduced by side branch classifiers are less than 2\% than the original ResNet-38 or MobileNetV2.

%\begin{figure}[ht]
%    \centering
%    \includegraphics[width=1\linewidth]{figures/resnet38v2.pdf}
%    \vspace{-1em}
%    \caption{Network architecture of RDI-ResNet38.}
%    \label{fig:res38v2}
%   \vspace{-1em}
%\end{figure}
%}
\section{Input-Adaptive Inference for RDI-Nets}\label{sec:6-3}	
Similar to the deterministic strategy in \cite{teerapittayanon2016branchynet}, we adopt the entropy as the measure of the prediction confidence. Given a prediction vector $y\in\bbR^{C}$, where $C$ is the number of class, the entropy of $y$ is defined as follow, 
\begin{equation}
-\sum_{c=1}^{C}(y_{c} + \epsilon)log(y_{c} + \epsilon),
\end{equation}
where $\epsilon$ is a small positive constant used for robust entropy computation. To perform fast inference on a ($K$+1)-output RDI-Net, we need to determine $K$ threshold numbers, \ie $\{t_{i}\}_{i=1}^{K}$, so that the input $x$ will exit at $i$th branch if the entropy of $y_{i}$ is larger than $t_{i}$. To choose $\{t_{i}\}_{i=1}^{K}$, \cite{huang2018multiscale} provides a good starting point  by fixing exiting probability of each branch classifiers equally on validation set so that each sample can equally contribute to inference. We follow this strategy but adjust the thresholds to make the contribution of middle branches slightly larger than the early branches. The threshold numbers for RDI-SmallCNN, RDI-ResNet38, and RDI-MobilenetV2 are set to be $\{0.023, 0.014\}$, $\{0.32, 0.36, 0.39, 0.83, 1.12, 1.35\}$, and $\{0.267, 0.765\}$, respectively.

\begin{figure}[ht]
    \centering
    %\vspace{-1em}
    \includegraphics[width=0.8\linewidth]{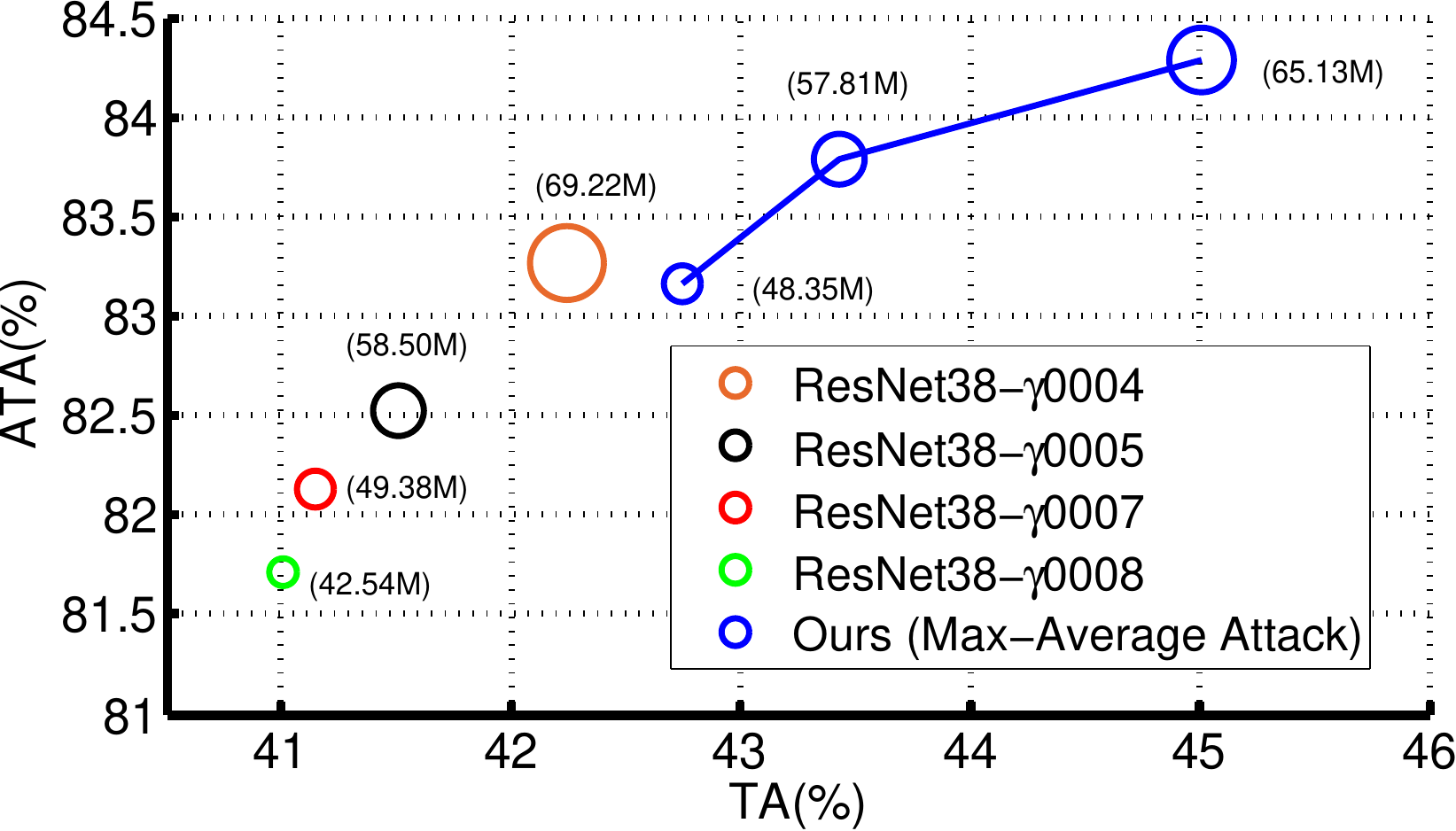}
    %vspace{-2em}
    \caption{Performance comparison between RDI-Net and the pruning + defense baseline. Each marker represents a model, whose size is proportional to its MFlops. $\gamma$ is the sparsity trade-off parameter: the larger the sparser (smaller model).
}    %\vspace{-2em}
    \label{fig:comparision}
\end{figure}

\section{Generalized Robustness}\label{sec:6-4}
Here, we introduce the attack form of random attack and report the complete results against FGSM \citep{goodfellow2014explaining} and WRM \citep{simha2018wrm} attacker under various attack forms, in Tables \ref{tab:complete_fgsm} and \ref{tab:complete_wrm}, respectively.

\paragraph{Random Attack} the attack exploits multi-loss flexibility by randomly fusing all $f_{\theta_{i}}$ losses. Given a $N$-output network, we have a fusion vector $C \in \bbR^{N} \sim \bbD$, where $\bbD$ is some distribution (uniform by default). We denote $c_{j}$ as the $j$th element of $C$ and $x^{adv}_{rdm}$ can be found by:
\begin{equation}
\hspace{-0.01in} x_{rdm}^{adv} = \argmax_{x'\in |x'-x|_{\infty}\leq\epsilon}|\frac{1}{N}\sum_{j=1}^{N}c_{j}\phi(f_{\theta_{j}}(x'), y)|.
\label{eqn:rdm-attack}
\end{equation}
%\textcolor{red}{Note that the sample space of $x^{adv}_{rdm}$ is infinite since $C$ is sampled from an infinite-sample distribution $\bbD$. 
It is expected to challenge our defense, due to the infinitely many ways of randomly fusing outputs.

\begin{table}[ht]
\centering
\vspace{-0.5em}
\caption{The performance evaluation on RDI-ResNet38 (defended with PGD) against FGSM attack. The perturbation size is $8/255$. The ATA of the original defended ResNet38 by PGD under the same attacker is $51.11\%$.}
\vspace{0.1in}
\label{tab:complete_fgsm}
\begin{tabular}{|l|l|l|l|l|}
\hline
Defence Method & Standard & Main Branch & Average & Max-Average \\ \hline
\hline
\textbf{TA} & 92.43\% & 83.74\% & 82.42\% & \textbf{83.79\%} \\ \hline
\hline
ATA (Branch1) & 20.69\% & 66.06\% & 72.77\% & 72.76\% \\ \hline
ATA (Branch2) & 16.15\% & 53.87\% & 70.40\% & 69.71\% \\ \hline
ATA (Branch3) & 8.13\% & 63.70\% & 64.19\% & 65.14\% \\ \hline
ATA (Branch4) & 10.09\% & 56.67\% & 58.45\% & 58.20\% \\ \hline
ATA (Branch5) & 9.45\% & 50.81\% & 52.76\% & 52.96\% \\ \hline
ATA (Branch6) & 10.22\% & 50.34\% & 53.17\% & 51.05\% \\ \hline
ATA (Main Branch) & 11.51\% & 51.45\% & 53.64\% & 54.72\% \\ \hline
ATA (Average) & 11.41\% & 50.21\% & 51.81\% & 53.20\% \\ \hline
ATA (Max-Average) & 2.09\% & 47.53\% & 50.63\% & 52.40\% \\ \hline
\textbf{ATA (Worst-Case)} & 2.09\% & 47.53\% & 50.63\% & \textbf{51.05\%} \\ \hline
\hline
Average MFlops & 65.74 & 55.27 & 58.27 & 59.67 \\ \hline
\textbf{Computation Saving} & 17.21\% & 30.40\% & 26.40\% & 24.86\% \\ \hline
\end{tabular}
\vspace{-0.5em}
\end{table}

\begin{table}[!ht]
\centering
\vspace{-0.5em}
\caption{The performance evaluation on RDI-ResNet38 (defended with PGD) against WRM attack. The perturbation size is $0.3$. The ATA of the original defended ResNet38 by PGD under the same attacker is $83.35\%$.}
\vspace{0.1in}
\label{tab:complete_wrm}
\begin{tabular}{|l|l|l|l|l|}
\hline
Defence Method & Standard & Main Branch & Average & Max-Average \\ \hline
\hline
\textbf{TA} & 92.43\% & 83.74\% & 82.42\% & \textbf{83.79\%} \\ \hline
\hline
ATA (Branch1) & 46.60\% & 83.73\% & 82.42\% & 83.78\% \\ \hline
ATA (Branch2) & 71.33\% & 83.73\% & 82.42\% & 83.79\% \\ \hline
ATA (Branch3) & 23.51\% & 83.73\% & 82.41\% & 83.78\% \\ \hline
ATA (Branch4) & 33.41\% & 83.73\% & 82.42\% & 83.78\% \\ \hline
ATA (Branch5) & 42.35\% & 83.73\% & 82.41\% & 83.78\% \\ \hline
ATA (Branch6) & 47.77\% & 83.74\% & 82.40\% & 83.78\% \\ \hline
ATA (Main Branch) & 34.42\% & 83.74\% & 82.42\% & 83.78\% \\ \hline
ATA (Average) & 26.48\% & 83.69\% & 82.36\% & 83.77\% \\ \hline
ATA (Max-Average) & 23.51\% & 83.73\% & 82.40\% & 83.78\% \\ \hline
\textbf{ATA (Worst-Case)} & 23.51\% & 83.69\% & 82.36\% & \textbf{83.77\%} \\ \hline
\hline
Average MFlops & 50.05 & 50.46 & 52.89 & 52.38 \\ \hline
\textbf{Computation Saving} & 36.98\% & 36.46\% & 33.40\% & 34.04\% \\ \hline
\end{tabular}
\vspace{-0.5em}
\end{table}

\end{document}